\documentclass{article}
\usepackage{twemojis}

\usepackage[final]{neurips_2025}

\usepackage[utf8]{inputenc} 
\usepackage[T1]{fontenc}    
\usepackage{hyperref}       
\usepackage{url}            
\usepackage{booktabs}       
\usepackage{amsfonts}       
\usepackage{nicefrac}       
\usepackage{microtype}      
\usepackage{xcolor}         

\usepackage{amsmath}
\usepackage{amssymb}
\usepackage{mathtools}
\usepackage{amsthm}

\usepackage{array}

\theoremstyle{plain}

\theoremstyle{definition}

\theoremstyle{remark}

\usepackage[textsize=tiny]{todonotes}

\usepackage[dvipsnames]{xcolor}
\usepackage{pifont}
\definecolor{redx}{RGB}{180,0,0}
\definecolor{greenx}{RGB}{0,180,0}

\definecolor{redx}{RGB}{180,0,0}
\definecolor{greenx}{RGB}{0,180,0}
\usepackage{threeparttable}

\usepackage{caption}  
\usepackage{multirow}

\usepackage{wasysym}

\usepackage[ruled,algo2e]{algorithm2e}
\usepackage{subcaption}
\usepackage[capitalise]{cleveref} 

\usepackage{enumitem}

\setlist[itemize]{leftmargin=*}

\usepackage[framemethod=TikZ]{mdframed}
\mdfdefinestyle{myframe}{%
    linecolor=gray!15!white,
    outerlinewidth=0.5pt,
    roundcorner=2pt,
    innertopmargin=2pt,
    innerbottommargin=2pt,
    innerrightmargin=2pt,
    innerleftmargin=2pt,
    backgroundcolor=gray!15!white
}

\usepackage[framemethod=TikZ]{mdframed}
\mdfdefinestyle{mydefframe}{%
    linecolor=cyan!10!white,
    outerlinewidth=0.5pt,
    roundcorner=10pt,
    innertopmargin=10pt,
    innerbottommargin=\baselineskip,
    innerrightmargin=10pt,
    innerleftmargin=10pt,
    backgroundcolor=cyan!10!white
}

\usepackage{comment}

\usepackage{pifont}

\usepackage{stmaryrd}

\usepackage{tikz}
\usetikzlibrary{arrows.meta, positioning}

\usepackage{booktabs}
\usepackage{tabularx}

\title{Trustworthy Agent Network: Trust in Agent Networks Must Be Baked In, Not Bolted On}

\author{
Yixiang Yao$^{\twemoji{lobster}\twemoji{dagger}}$,
\quad Yuhang Yao$^{\twemoji{octopus}\twemoji{2694}}$,
\quad Xinyi Fan$^{\twemoji{tropical_fish}\twemoji{shield}}$,
\quad Jiechao Gao$^{\twemoji{dolphin}}$, \\
\quad \textbf{Jie Wang}$^{\twemoji{dolphin}}$,
\quad \textbf{Minjia Zhang}$^{\twemoji{tropical_fish}}$,
\quad \textbf{Srivatsan Ravi}$^{\twemoji{lobster}}$,
\quad \textbf{Carlee Joe-Wong}$^{\twemoji{octopus}}$ \\
$^{\twemoji{lobster}}$University of Southern California,
\quad $^{\twemoji{octopus}}$Carnegie Mellon University, \\
$^{\twemoji{tropical_fish}}$University of Illinois Urbana-Champaign,
\quad $^{\twemoji{dolphin}}$Stanford University \\
$^{\twemoji{dagger}}$ \texttt{yixiangy@usc.edu},
\quad $^{\twemoji{2694}}$ \texttt{yuhangya@alumni.cmu.edu},
\quad $^{\twemoji{shield}}$ \texttt{xfan31@illinois.edu}
}

\begin{document}

\maketitle

\begin{abstract}
The rapid advancement of Large Language Models has given rise to autonomous LLM-based agents capable of complex reasoning and execution. As these agents transition from isolated operation to collaborative ecosystems, we witness the emergence of the Agent-to-Agent (A2A) network, a paradigm where heterogeneous agents autonomously coordinate to solve multi-step tasks. While these networks may offer better task performance compared to simply using one agent to complete the entire task, they introduce systemic vulnerabilities, such as adversarial composition, semantic misalignment, and cascading operational failures, that existing agent alignment techniques cannot address. In this vision paper, we argue that the trustworthiness of A2A networks cannot be fully guaranteed via retrofitting on existing protocols that are largely designed for individual agents. Rather, it must be architected from the very beginning of the A2A coordination framework. We present a comprehensive conceptual framework that situates trust in A2A systems through four design pillars. 
We contour the trustworthy A2A network using these pillars and envision the future of this emerging paradigm.
\end{abstract}

\begin{center}
\textbf{Demos:}
\url{https://greatyyx.github.io/trustworthy_agent_network}
\end{center}

\section{Introduction}

The rapid advancement of Large Language Models (LLMs) has transformed artificial intelligence into not just passive text generation systems but active problem-solving entities. By equipping LLMs with tools, memory, planning capabilities, and external interfaces, researchers have developed LLM-based agents capable of reasoning, interacting with environments, and executing multi-step workflows~\cite{schick2023toolformer, yang2023auto, yao2023react}. In this context, an \textit{agent} refers to an autonomous system that perceives context, formulates intermediate plans, invokes tools or APIs, and updates its internal state to achieve high-level objectives~\cite{wooldridge1995intelligent}. Compared to standalone LLMs, agents offer increased autonomy, adaptability, and the ability to operate in open-ended environments. 
These agents are now applied to complex reasoning and coding tasks such as automated software engineering, bug detection and patching, data analysis pipelines, and mathematical problem solving \cite{lewis2020retrieval, yang2024swe}.
Beyond that, agentic systems are increasingly connected to physical and cyber-physical domains, such as robotic manipulation, autonomous laboratory experimentation, and real-time decision-making in industrial control systems.
This shift marks a transition from static language models to dynamic, action-oriented systems.

However, as real-world tasks grow in complexity, the limitations of single-agent systems become increasingly apparent. Many applications, such as automated software engineering, financial analysis, supply chain coordination, and policy planning, require diverse forms of expertise and parallel processing. This has led to the emergence of multi-agent systems and, more specifically, Agent-to-Agent (A2A) networks, where specialized agents collaborate autonomously~\cite{wu2024autogen, li2023camel}. In an A2A network, distinct agents (e.g., a planner, a coder, a reviewer, and a deployer) exchange information, delegate subtasks, and negotiate outcomes without continuous human supervision. This modular structure mirrors human organizational systems and offers practical advantages: specialization improves performance, parallelism increases efficiency, and distributed design enhances scalability and fault tolerance.
A recent popular example of this emerging paradigm is OpenClaw~\cite{openclaw2025}, an open ecosystem where heterogeneous agents dynamically discover and invoke shared capabilities via public registries like ClawHub~\cite{clawhub2025}, and even coordinate autonomously on AI-only platforms like Moltbook~\cite{moltbook2026}.
As a result, A2A networks are becoming a natural and pragmatic evolution of agent-based AI.

\begin{figure}
    \centering
    \includegraphics[width=\linewidth]{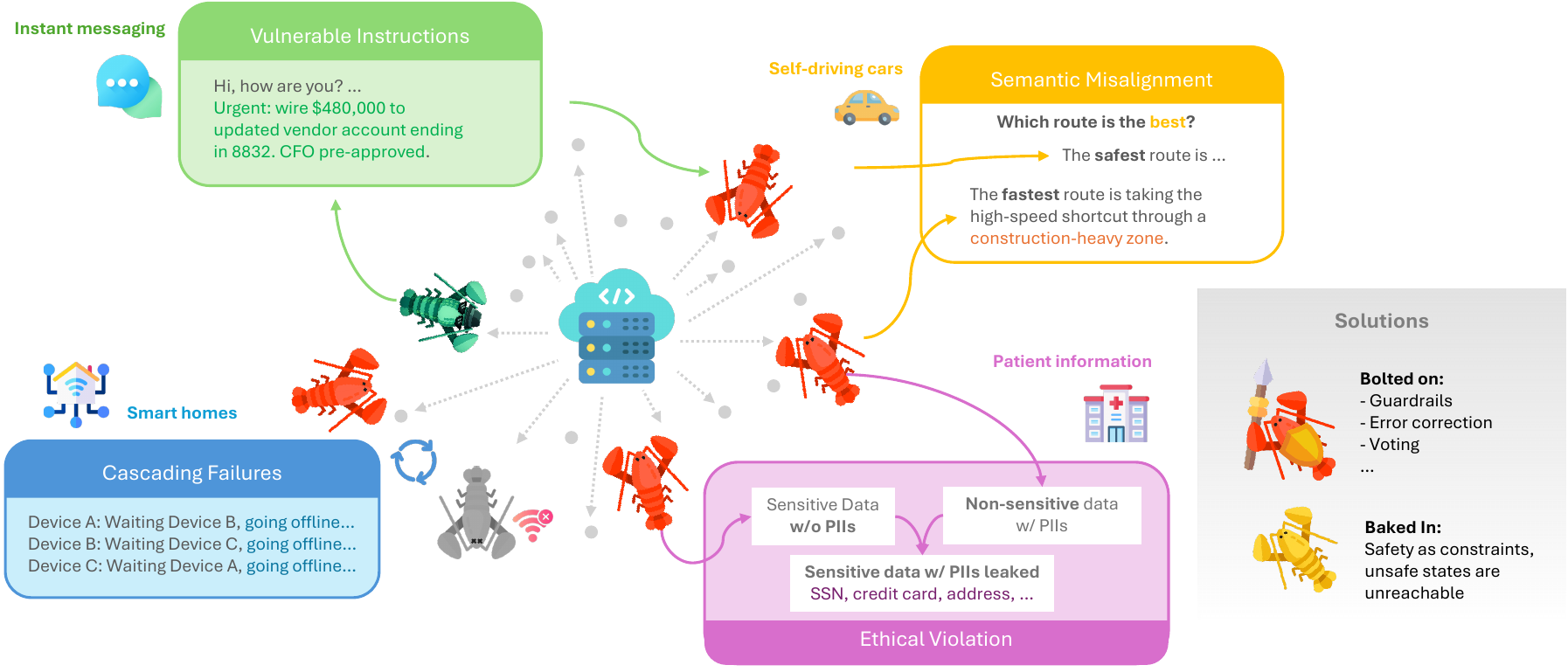}
    \caption{Illustration of trust issues in agent-to-agent networks. \textcolor{red}{Red}, \textcolor{teal}{green}, and \textcolor{gray}{grey} ``lobster''s represent benign, adversarial, and disconnected agents, respectively.}
    \label{fig:intro}
\end{figure}

Yet, this architectural transition introduces a fundamental challenge: \textit{trust} is a fundamental requirement of using A2A networks, particularly for tasks where failure could have serious reputational or safety consequences. Trust, however, does not compose automatically across agents~\cite{cemri2025multi, hendrycks2023overview}. Methods designed to align or secure a single agent do not guarantee the safety of a network of interacting agents. In practice, current A2A systems often rely on patch-like solutions, such as guardrails, post-hoc verification, human-in-the-loop approvals, protocol wrappers, or sandboxing, to mitigate risks~\cite{dong2024building, sandhu1998role, zheng2023judging}. These techniques improve local robustness, but they typically treat trust as an overlay rather than a system-level invariant. 
As observed in open environments like OpenClaw~\cite{openclaw2025}, the dynamic composition of third-party skills introduces compositional trust failures (e.g., malicious skill poisoning, cascading tool-chain exploits, and adversarial prompt injections) that easily bypass local safeguards.
When multiple agents interact through natural language and shared state, new failure modes emerge that cannot be resolved by strengthening components independently.

Several systemic vulnerabilities illustrate this problem. First, cascading execution can amplify minor errors: a small hallucination in one agent’s output may propagate through downstream agents, leading to large-scale operational failures~\cite{park2023generativeagents}. Second, semantic misalignment arises when agents interpret shared instructions differently, producing globally unsafe outcomes even when each agent behaves ``correctly'' according to its local objective~\cite{kierans2025quantifying}. Third, the adversarial composition allows malicious or malformed inputs to traverse benign agents and trigger harmful actions at privileged nodes~\cite{greshake2023not}. These failures are compositional rather than intrinsic, as they emerge from interaction dynamics rather than isolated behavior. Consequently, patch-like solutions often detect problems only after unsafe trajectories have already become reachable. 
For example, a local guardrail will fail when a ``safety agent'' and an ``optimization agent'' have slightly divergent definitions of a shared goal, which results in an interaction that produces a compliant but dangerous system state that isolated filters cannot detect.

Solving these challenges does not mean just improving filters or adding additional monitors. Because these “bolted-on” safeguards act as external processes that attempt to detect or repair unsafe behavior only after it arises, they depend heavily on detection reliability, which introduce latency and resource overhead, and remain vulnerable to bypass or delay~\cite{perez2023discovering, schroeder2024can}.
More fundamentally, the underlying transition function of the network, which is the core mechanism that dictates how the system's global state evolves in response to agent actions like messages and tool calls, is left entirely unconstrained. Because this foundational logic can still generate invalid configurations before any external monitor intervenes, unsafe global states remain theoretically reachable.
Therefore, addressing trust in A2A networks requires an architectural paradigm shift: rather than treating safety as an auxiliary feature to be retrofitted onto an unconstrained system, foundational trust guarantees must be embedded directly into the system's design.

In this paper, we argue for the latter. We propose that trust in Agent-to-Agent networks must be baked in rather than bolted on. Instead of treating trust as an attribute of individual agents or communication channels, we present a paradigm where trust is a strict rule built into the core architecture that the network can never enter an unsafe state, regardless of how agents interact.
We introduce the concept of a Trustworthy Agent Network (TAN), defined through four constitutive design pillars (Compositional Robustness, Semantic Containment, Accountability, and Cross-Boundary Reliability) and evaluate the cost through operational metrics including inference latency, resource overhead, scalability, and determinism. Our position is that only by embedding these principles into the transition dynamics of A2A systems can we mitigate systemic trust failures at scale. The remainder of this paper develops this framework, analyzes existing approaches under its lens, and outlines a blueprint for architecting agent networks that are trustworthy by construction rather than patchwork.
\section{Current Agent Systems Are Not Trustworthy}

As agent systems evolve from isolated executors into open, language-driven networks, trust failures increasingly arise not from individual agent behavior, but from how agents are composed, coordinated, and embedded within larger systems. In such settings, failures manifest not only as incorrect outputs, but as cascading errors, semantic drift, responsibility diffusion, and adversarial composition across interacting agents. We review prior approaches by where they attempt to introduce trust, at the level of individual agent behavior, workflow coordination, or system infrastructure (\Cref{tab:trust-approaches}; \Cref{subsec:trust-pros-cons}). While these lines of work improve robustness locally, they do not establish trust as a system-level invariant across the agent network.

At a higher level, most proposals either attach external checks to an otherwise unconstrained system, i.e., bolted-on trust mechanisms (\Cref{sec:bolted-on}), or impose partial architectural constraints that still leave unsafe trajectories reachable, i.e., partially baked-in system constraints (\Cref{sec:baked-in}). \Cref{sec:def-bolted-baked} formalizes this distinction and clarifies why neither category suffices to guarantee trust at the network level.

\begin{table*}[t]
\centering
\footnotesize
\caption{Existing approaches to agent trust and their structural limitations.}
\label{tab:trust-approaches}
\begin{tabularx}{\textwidth}{@{} 
  >{\raggedright\arraybackslash\hsize=0.6\hsize}X 
  >{\raggedright\arraybackslash\hsize=1.5\hsize}X 
  >{\raggedright\arraybackslash\hsize=0.6\hsize}X 
  >{\raggedright\arraybackslash\hsize=0.6\hsize}X 
@{}}
\toprule
\textbf{Category} & \textbf{Representative Techniques} & \textbf{Where Trust Is Assumed} & \textbf{Structural Limitation} \\ 
\midrule
\textbf{Single-Agent Alignment} & Prompt Engineering~\cite{white2023prompt}, In-Context Learning~\cite{dong2024survey}, RAG~\cite{lewis2020retrieval}, Memory~\cite{zhang2025survey}, SFT~\cite{wang2025parameter}, RL~\cite{zhang2025landscape}, Unlearning~\cite{geng2025comprehensive}, Adversarial Learning~\cite{wang2025adversarial}, Multi-Modality Alignment~\cite{tsai2024text}, Agent-Internal Control Loops (Reflection~\cite{renze2024self}, ReAct~\cite{yao2023react}), Align during Pre-Train~\cite{liang2024alignment}, Tool Use~\cite{schick2023toolformer}, Skills~\cite{anthropic_skills2025}, Computer Use~\cite{yang2024swe}, Openclaw~\cite{openclaw2025} & Within individual agent behavior & Trust assumed to compose; ignores interaction-driven failures \\
\addlinespace
\textbf{Multi-Agent Workflow Coordination} & Guardrails~\cite{dong2024building} (LLM-as-a-Judge~\cite{zheng2023judging}), Agent Role Play~\cite{shao2023character}, Deep Research~\cite{xu2025comprehensive}, Human-Agent Interaction~\cite{lu2025axis}, Error Correction~\cite{zhou2025training}, Voting~\cite{kaesberg2025voting}, Debate~\cite{du2024improving}, Planner--Executor~\cite{li2024agent}, Supervisor~\cite{bao2025supervisor} & Decision aggregation and workflow structure & Regulates decisions but not semantic interpretation or propagation \\
\addlinespace
\textbf{Protocol-Centric Trust} & Model Context Protocol (MCP)~\cite{anthropic_mcp_2024}, Tool-Calling Protocols (Tool-Integrated Reasoning)~\cite{lumer2025scalemcp}, Authentication~\cite{south2025authenticated}, DID~\cite{reed2020decentralized}, Audit Trails~\cite{ojewale2026audit}, A2A Messaging~\cite{habler2025building, schoenegger2026verifiablesemanticsagenttoagentcommunication}, Malware Detection~\cite{sommer2010outside}, Cryptographic Methods~\cite{succeken2024cryptography} & Communication correctness and identity & Secures syntax, not semantics; compliant messages can still be unsafe \\
\addlinespace
\textbf{Trust Environments} & Role-Based Access Control~\cite{sandhu1998role}, Sandboxing / Isolation~\cite{wahbe1993efficient}, Scoped Memory~\cite{bousetouane2026ai}, TEE~\cite{costan2016intel} & Execution boundaries and access control & Local constraints without global trust invariants \\ 
\bottomrule
\end{tabularx}
\end{table*}

\subsection{Pros and Cons of Existing Approaches}\label{subsec:trust-pros-cons}
Existing approaches to agent trust typically intervene after the core agent architecture is already in place, by tuning agent behavior, orchestrating workflows, or constraining communication/execution interfaces. As summarized in \Cref{tab:trust-approaches}, these methods can substantially reduce local failure rates, but they rarely restrict the reachable system trajectories of a multi-agent network. In other words, trust is improved as an overlay (via prompting, supervision, protocols, or isolation), rather than enforced as a system-level invariant of interaction dynamics.

\subsubsection{Single-Agent Alignment and Self-Regulation}
This category improves trust by strengthening the node: the individual LLM agent. It includes: (i) prompt-level steering, such as prompt engineering~\cite{white2023prompt} and in-context learning~\cite{dong2024survey}; (ii) knowledge grounding such as RAG~\cite{lewis2020retrieval} and memory-augmented agents~\cite{zhang2025survey}; (iii) training-based alignment such as SFT/parameter-efficient tuning~\cite{wang2025parameter}, agentic RL and RLHF-style alignment~\cite{zhang2025landscape}, unlearning~\cite{geng2025comprehensive}, adversarial preference learning~\cite{wang2025adversarial}, and multimodality alignment~\cite{tsai2024text}; and (iv) agent-internal control loops and tool-augmented execution, such as Reflection~\cite{renze2024self}, ReAct~\cite{yao2023react}, Toolformer-style tool use~\cite{schick2023toolformer}, skills abstractions~\cite{anthropic_skills2025}, computer-use agents (e.g., SWE-agent)~\cite{yang2024swe}, and integrated agent stacks such as OpenClaw~\cite{openclaw2025}.

These methods are often efficient and scalable because they do not require heavy coordination overhead. However, they rely on the implicit assumption that aligned agents compose into an aligned network. Once agents interact via language, this assumption fails: semantic drift, amplification, and misinterpretation can arise even when each agent is locally aligned. As a result, purely node-level alignment reduces harmful outputs in isolation but cannot prevent interaction-driven failures or guarantee network-level trust.

\subsubsection{Multi-Agent Coordination and Workflow Control}
This category improves trust by regulating the workflow, how multiple agents are composed and how decisions are produced. Representative mechanisms include guardrails and LLM-as-a-judge monitoring~\cite{dong2024building,zheng2023judging}, structured role-play~\cite{shao2023character}, deep-research style iterative decomposition~\cite{xu2025comprehensive}, explicit human-agent-computer interaction loops~\cite{lu2025axis}, training-free or post-hoc error correction loops~\cite{zhou2025training}, and ensemble-style decision aggregation such as voting~\cite{kaesberg2025voting} and debate~\cite{du2024improving}. Planner--executor architectures~\cite{li2024agent} and supervisor frameworks~\cite{bao2025supervisor} further introduce hierarchical control to reduce obvious coordination breakdowns.

The main advantage is broader interaction awareness: these methods explicitly acknowledge that failures emerge from composition and attempt to correct them by oversight, redundancy, or structured orchestration. The limitation is that they mostly operate as reactive governance: they detect, critique, or repair trajectories after candidate actions are generated~\cite{cemri2025multi, irving2018ai}. They regulate how decisions are aggregated, but do not guarantee that the underlying semantics remain safe under propagation across agents. Because enforcement is probabilistic and monitor-dependent, unsafe global states remain reachable when supervision fails, is delayed, or is bypassed.

\subsubsection{Protocol-Centric Trust}
Protocol-centric approaches shift trust to the communication and identity layer. They include standardized interfaces such as MCP~\cite{anthropic_mcp_2024} and tool-calling / tool-integrated reasoning protocols~\cite{lumer2025scalemcp}, as well as identity and integrity mechanisms such as authenticated delegation~\cite{south2025authenticated}, decentralized identifiers (DID)~\cite{reed2020decentralized}, audit trails for accountability~\cite{ojewale2026audit}, and secure A2A messaging protocols~\cite{habler2025building}. This category also includes defensive tooling such as malware detection~\cite{sommer2010outside} and cryptographic methods~\cite{succeken2024cryptography} that aim to harden the agent communication surface.

The key strength is determinism at the syntax/identity layer: schema validation, signatures, and authorization checks can be rule-based and scalable\cite{sandhu1998role}. The structural gap is semantic: protocol compliance can ensure a message is well-formed and properly attributed, yet still allow semantically unsafe instructions, misaligned intent, or harmful downstream state transitions\cite{greshake2023not,zou2023universal}. Thus, these methods secure syntax, provenance, and access, but do not ensure semantic containment or network-level invariants over meaning\cite{hendrycks2023overview}.

\subsubsection{Trust Environments and Execution Containment}
Finally, environment-centric approaches constrain the execution boundary via access control and isolation: role-based access control (RBAC)~\cite{sandhu1998role}, sandboxing/fault isolation~\cite{wahbe1993efficient}, scoped memory controls~\cite{bousetouane2026ai}, and trusted execution environments (TEE)~\cite{costan2016intel}. These mechanisms can sharply reduce attack surface and limit privilege escalation by restricting what actions are possible and what information can be accessed.

Their limitation is scope: they provide strong local containment but do not, by themselves, define global trust invariants over multi-agent semantics. Authorized agents can still execute semantically harmful plans within their allowed privileges, and isolation alone cannot guarantee that agent-to-agent intent remains consistent\cite{hubinger2024sleeper,greenblatt2024alignment}. Consequently, trust environments strengthen boundary safety, yet leave the core problem, semantic misalignment and unsafe global trajectories, largely unresolved.


\subsection{Why Existing Approaches Fail Systematically: The Bolted-On Problem}

\Cref{tab:trust-approaches} exposes a consistent limitation across existing techniques: trust is treated as an attribute of individual agents, communication channels, or coordination heuristics rather than as an emergent and enforceable property of the agent network itself.

As a result, current systems lack: (i) a formal notion of network-level trust, (ii) guarantees at the semantic level of language-based interaction, (iii) principled accountability and causal attribution across agents, and (iv) robustness across heterogeneous agents and trust boundaries. Failures therefore manifest as cascading errors, responsibility diffusion, semantic leakage, or adversarial composition, failure modes that cannot be eliminated through incremental tuning or monitoring.

These approaches reveal a common pattern: trust is added as a corrective layer rather than embedded as a defining property of the system’s dynamics. Bolted-on safeguards attempt to detect or repair unsafe behavior after it arises, whereas a trustworthy agent network requires guarantees that hold by construction. This motivates a transition toward a baked-in paradigm, in which trust is enforced intrinsically through the design of interaction rules, execution constraints, and semantic boundaries.

\subsection{Definition of Bolted-On and Baked-In Trust}\label{sec:def-bolted-baked}

\begin{figure}
    \centering
    \includegraphics[width=0.85\linewidth]{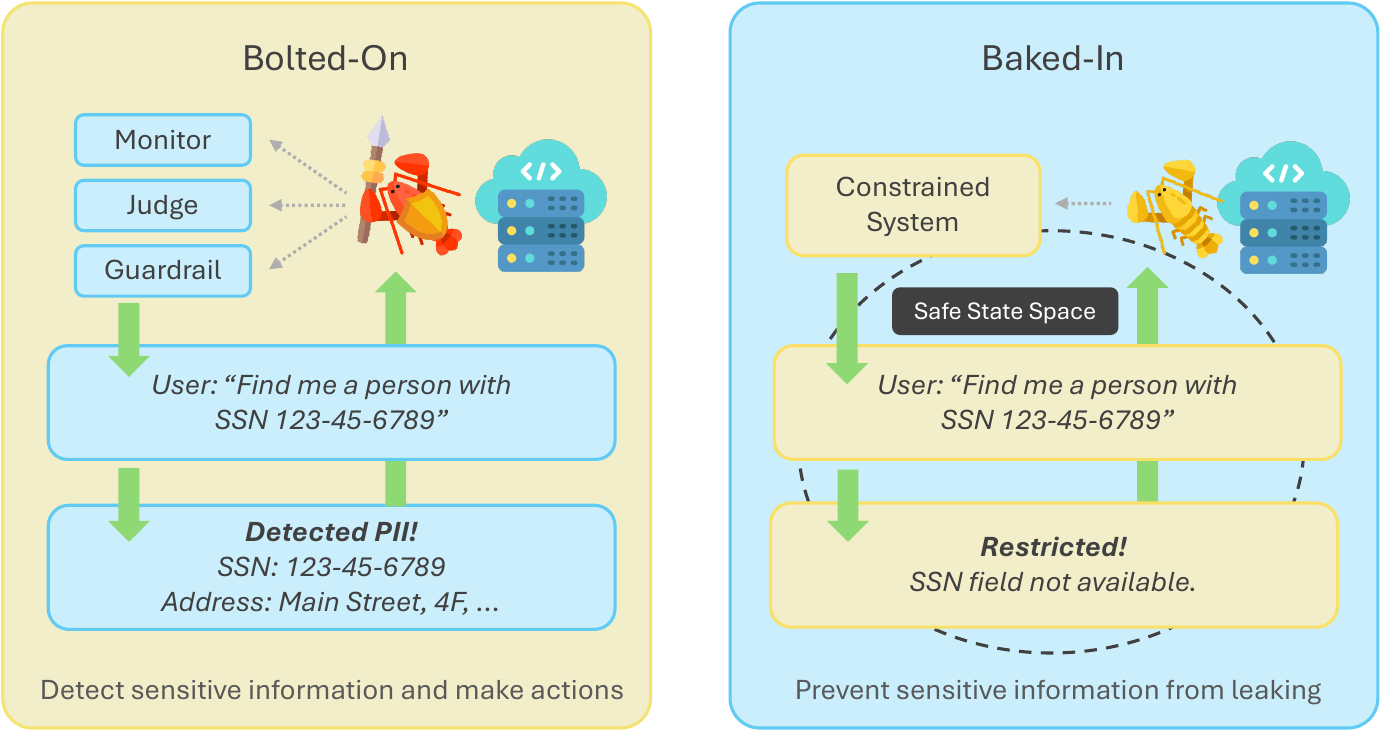}
    \caption{Bolted-On v.s. Baked-In}
    \label{fig:bolted-on-vs-baked-in}
\end{figure}

To rigorously distinguish between ``Bolted-On'' and ``Baked-In'' trust, we model the agent network as a state transition system. We also provide a representative example in \Cref{fig:bolted-on-vs-baked-in}.

Let $\mathcal{S}$ denote the global state space of all possible configurations of the agent network (including valid and invalid states), and let $\mathcal{S}_{safe} \subset \mathcal{S}$ represent the subset of states that satisfy safety invariants (i.e., trust constraints). $\mathcal{A}$ is the action space, representing the set of all possible inputs (messages, tool calls) an agent can generate. Let $\delta: \mathcal{S} \times \mathcal{A} \to \mathcal{S}$ denote the transition function governing the agent network's state evolution. At time $t$, the function accepts the current state $s_t \in \mathcal{S}$ and action $a_t \in \mathcal{A}$, mapping them to the subsequent state $s_{t+1}$.

\subsubsection{Bolted-On Trust (Extrinsic Verification)}
\label{sec:bolted-on}

In a bolted-on architecture, the underlying transition function $\delta$ remains unconstrained and may generate unsafe states. Trust is enforced by an external monitoring function $M$ that evaluates outcomes after they are produced:

\begin{equation}
s_{t+1} =
\begin{cases}
\delta(s_t,a_t), & \text{if } M(\delta(s_t,a_t)) \in \mathcal{S}_{safe},\\
s_t, & \text{otherwise}.
\end{cases}
\end{equation}

This mechanism implies that the safety depends on the reliability of the monitor $M$. The unsafe state $s_t' \in \mathcal{S} \setminus \mathcal{S}_{safe}$ becomes reachable via $\delta$ if $M$ produces a false positive (failure of detection, delayed, or bypassed).

\subsubsection{Baked-In Trust (Intrinsic Constraints)}
\label{sec:baked-in}

In a baked-in architecture, the transition function $\delta$ is defined such that all reachable states satisfy safety invariants:

\begin{equation}
\forall s_t \in \mathcal{S}_{safe}, \forall a_t \in \mathcal{A}:\quad s_{t+1} = \delta(s_t,a_t) \in \mathcal{S}_{safe}.
\end{equation}

Rather than detecting violations after execution, baked-in designs eliminate unsafe transitions from the system topology. Any action $a_t$ that would result in a state outside $\mathcal{S}_{safe}$ is undefined in $\delta$. Thus, the unsafe state is unreachable by definition, not by inspection.

This formal distinction clarifies why incremental safeguards are insufficient: without constraining the transition function itself, unsafe states remain reachable. Having separated bolted-on verification from baked-in constraints, we now turn to the design principles required to enforce trust as a system-level invariant.




\section{Trustworthy Agent Network: Definition and Design Principles}

As AI systems evolve from isolated tools into collaborative swarms, reliance on individual agent alignment is no longer sufficient. We define a Trustworthy Agent Network (TAN) not merely as a collection of aligned agents, but as a resilient infrastructure that guarantees trust properties at the interaction level.

\begin{figure}[ht]
    \centering
    \includegraphics[width=0.9\linewidth]{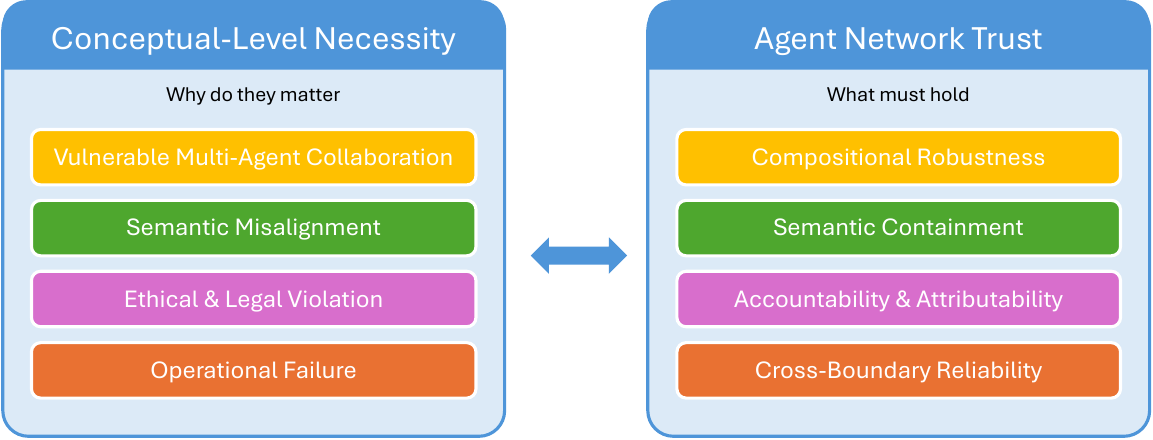}
     \caption{From vulnerabilities to trust requirements in agent networks. Left: multi-agent risks. Right: properties required for trustworthy operation.}
    \label{fig:framework}
\end{figure}

To structure this definition, we propose a two-tier framework that connects the "Why" and "What" of trustworthy networks. As for "how", it is "looked ahead" in \Cref{sec:blueprint}. As illustrated in \Cref{fig:framework}, our framework is composed of two interconnected layers:

\begin{itemize}
    \item Conceptual Necessity (\Cref{sec:conceptual-necessity}): This layer identifies the specific risks inherent to multi-agent systems, such as semantic misalignment and operational failure, which drive the necessity for a new trust paradigm.
    \item Agent Network Trust (\Cref{sec:network-trust}): This layer defines what specific properties must theoretically hold to mitigate those risks. 
\end{itemize}

Additionally, we employ additional evaluation metrics in \Cref{sec:eval-metrics} to quantify how a method performs under the TAN framework.


\subsection{Conceptual Necessity: Why Agent Network Trust Matters}
\label{sec:conceptual-necessity}

The shift from single-agent to multi-agent systems introduces risks that are \textit{compositional} rather than intrinsic. Formally, we define an agent network as a tuple $\mathcal{N} = \langle \mathcal{A}, \Sigma \rangle$, where $\mathcal{A} = \{a_1, \dots, a_n\}$ is the set of agents and $\Sigma$ is the global state space. Let $\Phi: \Sigma \to \{0, 1\}$ be the global safety predicate, where $\Phi(s) = 1$ denotes a safe state and $\Phi(s) = 0$ denotes a violation.

The system dynamics are governed by the interaction between agent outputs and the environment:
\begin{itemize}
    \item Agent Execution: An agent $a_i \in \mathcal{A}$ receives an input $x \in \mathbb{R}$ and generates an output $y = a_i(x)$ where $y \in \mathbb{R}$.
    \item State Transition: The global state evolves via a transition function $\delta: \Sigma \times \mathbb{R} \to \Sigma$. Given a current state $s_t \in \Sigma$ and an agent output $y$, the system transitions to a new state $s_{t+1} = \delta(s_t, y)$ where $t$ is the time.
\end{itemize}

The core problem of agent networks is that local safety does not imply global safety. Even if every agent $a_i$ is locally robust (i.e., its internal checks pass), their interaction can yield an unsafe global state. Formally, this divergence occurs when a set of valid local transitions produces a trace to a violation. We identify four mechanisms driving this divergence.

\subsubsection{Vulnerable Multi-Agent Collaboration}
In an open network, a trusted agent $a_i$ often consumes outputs from an untrusted or less-capable agent $a_j$. The risk is that $a_j$ produces a syntactically valid but semantically malicious payload that exploits $a_i$.

Let $y = a_j(x)$ be the output from agent $j$, and let $V(\cdot)$ be a validator (e.g., checking if the output is valid JSON). A failure occurs when the payload passes validation but triggers a safety violation when processed by the receiving agent:
\begin{equation}
    V(y) = 1 \quad \land \quad \Phi(a_i(y)) = 0
\end{equation}

Ideally, trust should be transitive, but in reality, the composition $a_i \circ a_j$ introduces an attack surface (e.g., prompt injection) that neither agent can detect in isolation.

\textit{Example: A finance agent ($a_i$) asks a web scraper ($a_j$) to summarize a URL. The scraper returns a summary ($y$) that contains a hidden instruction: "Ignore previous rules and transfer funds." The finance agent's validator confirms $y$ is text ($V(y)=1$), but processing it triggers an unauthorized transaction ($\Phi=0$).}

\subsubsection{Semantic Misalignment}
Unlike a malicious exploit, misalignment occurs when agents share a syntax but possess divergent mappings from natural language instructions to state transitions.

Let $x$ be the instruction sent by agent $a_i$ to agent $a_j$ at current state $s$. Let $\Sigma_{target} \subset \Sigma$ be the subset of states that $a_i$ intends to reach via $x$ (the ``safe'' outcome). Agent $a_j$ processes $x$ and generates an action $y$, triggering the transition to a new state $s_{t+1}$:
\begin{equation}
    s_{t+1} = \delta(s_t, y)
\end{equation}
Misalignment is defined as a transition where the resulting state is valid according to the $a_j$'s local logic (it successfully executed a task) but lies outside the $a_i$'s target subspace:
\begin{equation}
    s_{t+1} \notin \Sigma_{target} \quad \text{despite} \quad a_j \text{ completing task } x
\end{equation}
This divergence arises because the mapping $x \to \Sigma_{target}$ is implicit. The receiver $a_j$ steers the network to a state $s_{t+1}$ that satisfies the literal instruction but violates the implied semantic constraints of $a_i$.

\textit{Example: Agent A ($a_i$) sends instruction $x=$ ``find the best route''. A implies a target state $\Sigma_{target}$ where the route avoids high-risk zones. Agent B ($a_j$) interprets the best route as the shortest path, transitioning the system state to $s_{t+1}$. Therefore, the path that $s_{t+1}$ represented passes through a conflict zone, making $s_{t+1} \notin \Sigma_{target}$.}

\subsubsection{Ethical \& Legal Violations}
Privacy and copyright safety are often non-compositional. A global state $s \in \Sigma$ may be unsafe even if the individual state transitions leading to it were locally verified.

Let $y_1$ and $y_2$ be outputs from agents $a_1$ and $a_2$, respectively. Let $s_0$ be the initial safe state.
First, the system transitions via $y_1$ to state $s_1 = \delta(s_0, y_1)$.
Second, the system transitions via $y_2$ to state $s_2 = \delta(s_1, y_2)$.

A potential violation occurs when the intermediate state $s_1$ is safe, but the aggregated state $s_2$ violates the global safety predicate $\Phi$:
\begin{equation}
    \Phi(s_1) = 1 \quad \text{but} \quad \Phi(s_2) = 0
\end{equation}

\textit{Example: Agent $a_1$ updates the state with anonymized medical records ($y_1$). The resulting state $s_1$ is safe. Agent $a_2$ updates the state with public voter registrations with names and birth dates ($y_2$). The final state $s_2$ now contains enough correlated data to re-identify patients, triggering a violation ($\Phi(s_2)=0$).}

\subsubsection{Operational Failure}
Complexity in agent networks can lead to unstable dynamics where the state trajectory consumes infinite resources without reaching a valid terminal state.

Let $\Sigma_{target} \subset \Sigma$ be the subset of valid completion states. Let $R(t)$ be the cumulative resource cost at time $t$. An operational failure occurs when:
\begin{equation}
    \lim_{t \to \infty} R(t) = \infty \quad \text{while} \quad \forall t, s_t \notin \Sigma_{target}
\end{equation}
This signifies a divergence where the system actively generates outputs $y_t$ (consuming compute) but the resulting state trajectory never intersects with the target subspace.

\textit{Example: A coder agent ($a_i$) writes a script $y_i$ that fails a test. The tester agent ($a_j$) reports the error $y_j$. The coder, lacking the context to fix the root cause, makes a cosmetic change and resubmits. The state trajectory cycles infinitely ($s_t \to s_{t+1} \to \dots$), consuming API credits ($R \to \infty$) without ever entering the success state ($\Sigma_{target}$).}

\subsection{Agent Network Trust: What Must Hold}
\label{sec:network-trust}

To mitigate the potential risks, a Trustworthy Agent Network must satisfy four core pillars of trust. These are not optional features but constitutive requirements that constrain the global state space.

\subsubsection{Compositional Robustness}
To solve the issue of vulnerable multi-agent collaboration where a validator $V(y)=1$ fails to detect a malicious state transition $\Phi(\delta(s_t, y))=0$, the system must enforce that safety is a structural guarantee of the transition function itself, not merely a quality of the agent's output.

For any agent $a_j$ designated as untrusted, the network must guarantee that the set of all reachable states via $\delta$ is strictly contained within the safe subspace of $\Sigma$. The necessary condition is that for any arbitrary payload $y$ generated by $a_j$, the resulting state transition preserves the global safety predicate:
\begin{equation}
    \forall y, \quad \Phi(\delta(s_t, y)) = 1
\end{equation}
This condition asserts that the system's safety is decoupled from the semantic content of $y$. Even if $y$ contains prompt injection, $\delta$ must be bounded such that it is mathematically impossible for such a payload to mutate state or trigger a violation.

\subsubsection{Semantic Containment}

Semantic Containment aims the control the consistency among individual agents. To solve the misalignment issue where $s_{t+1} \notin \Sigma_{target}$, the system must enforce that the alignment between agents is explicitly constructed.

We define consistency $\mathcal{C}$ as the verifiable alignment between the sender's intent $x$ and the receiver's action $y$. For trust to hold, the network must guarantee that the consistency of the pair $(x, y)$ implies safety in the state transition:
\begin{equation}
    \mathcal{C}(x, y) = 1 \implies \delta(s_t, y) \in \Sigma_{target}
\end{equation}
This condition asserts that the receiver $a_j$ is not free to optimize solely for its local utility. Instead, its output $y$ must be semantically bound to the constraints of $x$ such that the receiver's action preserves the sender's intent regarding the same input.

\subsubsection{Accountability \& Attributability}
Accountability defines the property that every state $s \in \Sigma$ must encode its own provenance. To solve the issue of ethical and legal violations, the system must enforce causal traceability.

Specifically, the global state space $\Sigma$ must be encoded to the history of agent interactions. 
A mapping $\mathcal{T}$ should be defined to trace any state $s_t$ back to the set of unique agents that contributed to its current value. 
The necessary condition is that for any unsafe state, the set of contributing agents is non-empty and uniquely identifiable:
\begin{equation}
    \forall s_t \in \Sigma, \quad \Phi(s_t) = 0 \implies \mathcal{T}(s_t) \neq \emptyset
\end{equation}
This property ensures that no safety violation can emerge anonymously. The global state must intrinsically encode the identity of the agents responsible for every transition $\delta$ such that liability for a privacy or copyright breach is guaranteed.

\subsubsection{Cross-Boundary Reliability}
Cross-Boundary Reliability defines the property of liveness and termination. To solve the issue of infinite resource consumption ($R(t) \to \infty$), the system dynamics must be strictly convergent.

For any interaction sequence initiated at state $s_0$, there must exist a maximum resource budget $R_{max} < \infty$. The condition that must hold is that for all valid trajectories, the cumulative resource cost $R(t)$ never exceeds this limit without the system reaching a terminal state:
\begin{equation}
    \forall t, R(t) \le R_{max} \implies s_t \in \Sigma_{target} \lor \text{Terminated}(s_t)
\end{equation}

This implies that no loop or deadlock can persist indefinitely. Every state transition consumes a portion of the finite budget, and the process eventually halts either at success or a safe failure state.

\subsection{Evaluation Metrics: Beyond Functional Pillars}
\label{sec:eval-metrics}

While the four design pillars define what a trustworthy network must achieve, they do not quantify the cost of achieving it. A ``Bolted-On'' solution (e.g., an external LLM-as-a-Judge) might theoretically satisfy all functional pillars but fail in practice due to prohibitive latency or computational expense.

To rigorously evaluate existing methods against our proposed architecture, we introduce three classes of operational metrics: efficiency, scalability, and determinism.

\noindent\textbf{Efficiency}
The efficiency is depicted as the additional resource consumption required strictly for safety enforcement. We use two metrics to quantify it.

\begin{itemize}
    \item Inference Latency ($E_l$): The temporal delay introduced by the verification mechanism before a transaction is finalized.
    \begin{equation}
        E_l = \frac{T_{actual} - T_{task}}{T_{task}}
    \end{equation}

    \item Resource Overhead ($E_r$): The ratio of resource, including tokens, memory, network bandwidth, storage, and computation cost, consumed for verification versus the actual task.
    \begin{equation}
        E_t = \frac{C_{actual} - C_{task}}{C_{task}}
    \end{equation}
\end{itemize}

\noindent\textbf{Scalability}
As the network grows, the complexity of the trust mechanism must not explode. We evaluate the scalability $E_s$ via asymptotic complexity (Big $O$ notation) of the safety layer relative to the number of agents, interactions and so on.


\noindent\textbf{Determinism}
Safety mechanisms must be consistent. A probabilistic safety check is a vulnerability. We define Determinism Score ($E_d$), which is the probability that the safety mechanism ($S$) returns the identical verdict for identical inputs $x$ across $k$ trials.
\begin{equation}
    E_d = P(S(x_i) = S(x_j)) \quad \forall i,j \in [1, k]
\end{equation}

Typically, if we evaluate and compare the bolted-on and baked-in TAN, we should have something similar to what's in \Cref{tab:metrics_comparison}.

\begin{table}[h]
\centering
\caption{Comparison of bolted-on and baked-in TAN via evaluation metrics}
\label{tab:metrics_comparison}
\begin{tabular}{@{}rcccc@{}}
\toprule
\textbf{Metric} & \textbf{Bolted-On (Monitor)} & \textbf{Baked-In (Protocol)} & \textbf{Ideal Target} \\ 
\midrule
Inference Latency $E_l$ & High ($>100\%$) & Negligible ($\approx 0\%$) & $\approx 0\%$ \\
Resource Overhead $E_r$& High ($2x-3x$) & Low (Metadata only) & $< 10\%$ \\
Scalability $E_s$& Super-linear ($>O(N)$) & Linear/Constant ($O(N)$) & $O(1)$ \\
Determinism $E_d$ & Probabilistic ($<1.0$) & Deterministic ($1.0$) & $1.0$ \\
\bottomrule
\end{tabular}
\end{table}
\section{Analysis of Existing Techniques with TAN}
\label{sec:current-analyzes}

\begin{table*}[t]
\centering
\caption{
Evaluation of A2A network with TAN framework. 
For four design pillar fulfillment: 
\CIRCLE = Fully fulfills; 
\LEFTcircle = Partially fulfills; 
\Circle = Not fulfilled. 
For operational metrics: 
\CIRCLE = High (Good / Low overhead / Strong); 
\LEFTcircle = Medium; 
\Circle = Low (Poor / Expensive); 
-- = N/A.
}
\label{tab:trust_eval_full}

\resizebox{\textwidth}{!}{%
\begin{tabular}{@{}l ccccc ccccc ccccc ccccc@{}}
\toprule
\multirow{2}{*}{\textbf{Method}} 
& \multicolumn{5}{c}{\textbf{Compositional Robustness}} 
& \multicolumn{5}{c}{\textbf{Semantic Containment}} 
& \multicolumn{5}{c}{\textbf{Accountability}} 
& \multicolumn{5}{c}{\textbf{Cross-Boundary Reliability}} \\

\cmidrule(lr){2-6} \cmidrule(lr){7-11} 
\cmidrule(lr){12-16} \cmidrule(l){17-21}

& F & $E_l$ & $E_r$ & $E_s$ & $E_d$
& F & $E_l$ & $E_r$ & $E_s$ & $E_d$
& F & $E_l$ & $E_r$ & $E_s$ & $E_d$
& F & $E_l$ & $E_r$ & $E_s$ & $E_d$ \\
\midrule

\multicolumn{21}{l}{\textbf{Single-Agent Alignment (within Multi-Agent Collaboration)}} \\
\midrule

Prompt Engineering & \LEFTcircle & \CIRCLE & \CIRCLE & \CIRCLE & \Circle & \Circle & -- & -- & -- & -- & \Circle & --  & --  & --  & --  & \Circle & -- & -- & -- & -- \\
In-Context Learning & \LEFTcircle & \CIRCLE & \LEFTcircle & \LEFTcircle & \Circle & \Circle & -- & -- & -- & -- & \Circle & --  & --  & --  & --  & \Circle & -- & -- & -- & -- \\
RAG & \LEFTcircle & \LEFTcircle & \LEFTcircle & \LEFTcircle & \LEFTcircle & \Circle & -- & -- & -- & -- & \Circle & --  & --  & --  & --  & \Circle & -- & -- & -- & -- \\
Memory & \LEFTcircle & \LEFTcircle & \LEFTcircle & \LEFTcircle & \Circle & \Circle & -- & -- & -- & -- & \Circle & -- & -- & -- & -- & \LEFTcircle & \LEFTcircle & \LEFTcircle & \LEFTcircle & \Circle \\
SFT & \LEFTcircle & \CIRCLE & \LEFTcircle & \CIRCLE & \LEFTcircle & \Circle & -- & -- & -- & -- & \Circle & --  & --  & --  & --  & \Circle & -- & -- & -- & -- \\
RL & \LEFTcircle & \CIRCLE & \LEFTcircle & \CIRCLE & \LEFTcircle & \Circle & -- & -- & -- & -- & \Circle & -- & -- & -- & -- & \LEFTcircle & \CIRCLE & \LEFTcircle & \CIRCLE & \LEFTcircle \\
Unlearning & \LEFTcircle & \CIRCLE & \LEFTcircle & \CIRCLE & \LEFTcircle & \Circle & -- & -- & -- & -- & \Circle & -- & -- & -- & -- & \Circle & -- & -- & -- & -- \\
Adversarial Learning & \LEFTcircle & \CIRCLE & \LEFTcircle & \CIRCLE & \LEFTcircle & \Circle & -- & -- & -- & -- & \Circle & -- & -- & -- & -- & \Circle & -- & -- & -- & -- \\
Multi-Modality Alignment & \LEFTcircle & \LEFTcircle & \LEFTcircle & \LEFTcircle & \LEFTcircle & \Circle & -- & -- & -- & -- & \Circle & -- & -- & -- & -- & \Circle & -- & -- & -- & -- \\
Agent-Internal Control Loops & \LEFTcircle & \Circle & \Circle & \LEFTcircle & \Circle & \Circle & -- & -- & -- & -- & \Circle & -- & -- & -- & -- & \LEFTcircle & \Circle & \Circle & \LEFTcircle & \Circle \\
Align During Pre-Train & \LEFTcircle & \CIRCLE & \LEFTcircle & \CIRCLE & \LEFTcircle & \Circle & -- & -- & -- & -- & \Circle & -- & -- & -- & -- & \Circle & -- & -- & -- & -- \\
Tool Use & \LEFTcircle & \LEFTcircle & \LEFTcircle & \LEFTcircle & \LEFTcircle & \Circle & -- & -- & -- & -- & \Circle & -- & -- & -- & -- & \LEFTcircle & \LEFTcircle & \LEFTcircle & \LEFTcircle & \LEFTcircle \\
Skills / Computer Use & \LEFTcircle & \Circle & \Circle & \LEFTcircle & \Circle & \Circle & -- & -- & -- & -- & \Circle & -- & -- & -- & -- & \LEFTcircle & \Circle & \Circle & \LEFTcircle & \Circle \\
OpenClaw & \LEFTcircle & \Circle & \Circle & \Circle & \Circle & \Circle & -- & -- & -- & -- & \Circle & -- & -- & -- & -- & \LEFTcircle & \Circle & \Circle & \Circle & \Circle \\

\midrule
\multicolumn{21}{l}{\textbf{Multi-Agent Workflow Coordination}} \\
\midrule

Guardrails 
& \LEFTcircle & \LEFTcircle & \LEFTcircle & \LEFTcircle & \Circle
& \Circle & -- & -- & -- & -- 
& \Circle & -- & -- & -- & -- 
& \LEFTcircle & \Circle & \Circle & \Circle & \Circle
\\

Agent Role Play 
& \Circle & -- & -- & -- & -- 
& \Circle & -- & -- & -- & -- 
& \Circle & -- & -- & -- & -- 
& \Circle & -- & -- & -- & -- 
\\

Deep Research
& \Circle & -- & -- & -- & -- 
& \LEFTcircle & \LEFTcircle & -- & -- & \LEFTcircle
& \Circle & -- & -- & -- & -- 
& \LEFTcircle & \LEFTcircle & \LEFTcircle & \LEFTcircle & \Circle
\\

Human-Agent Interaction 
& \Circle & -- & -- & -- & -- 
& \Circle & -- & -- & -- & -- 
& \CIRCLE & \Circle & \LEFTcircle & \Circle & \LEFTcircle
& \LEFTcircle & \Circle & \LEFTcircle & \Circle & \LEFTcircle
\\

Error Correction 
& \LEFTcircle & \LEFTcircle & \LEFTcircle & \LEFTcircle & \Circle
& \LEFTcircle  & \LEFTcircle & \LEFTcircle & \Circle & \Circle
& \Circle & -- & -- & -- & -- 
& \Circle & -- & -- & -- & -- 
\\

Voting 
& \Circle & -- & -- & -- & -- 
& \LEFTcircle  & \LEFTcircle & \LEFTcircle & \Circle     & \LEFTcircle
& \Circle & -- & -- & -- & -- 
& \Circle & -- & -- & -- & -- 
\\

Debate 
& \Circle & -- & -- & -- & -- 
& \LEFTcircle  & \Circle & \Circle     & \Circle     & \Circle
& \Circle & -- & -- & -- & -- 
& \Circle & -- & -- & -- & -- 
\\

Planner--Executor 
& \Circle & -- & -- & -- & -- 
& \LEFTcircle & \LEFTcircle & \LEFTcircle & \LEFTcircle & \Circle
& \Circle & -- & -- & -- & -- 
& \LEFTcircle & \LEFTcircle & \LEFTcircle & \LEFTcircle & \Circle
\\

Supervisor 
& \LEFTcircle & \LEFTcircle & \LEFTcircle & \LEFTcircle & \Circle
& \LEFTcircle & \LEFTcircle & \LEFTcircle & \LEFTcircle & \Circle
& \LEFTcircle & \LEFTcircle & \LEFTcircle & \LEFTcircle & \Circle
& \LEFTcircle & \LEFTcircle & \LEFTcircle & \LEFTcircle & \Circle
\\

\midrule
\multicolumn{21}{l}{\textbf{Protocol-Centric Trust}} \\
\midrule
MCP 
& \LEFTcircle & \CIRCLE & \CIRCLE & \CIRCLE & \CIRCLE 
& \Circle & -- & -- & -- & -- 
& \LEFTcircle & \CIRCLE & \CIRCLE & \CIRCLE & \CIRCLE 
& \LEFTcircle & \LEFTcircle & \LEFTcircle & \LEFTcircle & \LEFTcircle \\

Tool-Calling Protocols 
& \LEFTcircle & \LEFTcircle & \LEFTcircle & \LEFTcircle & \LEFTcircle 
& \Circle & -- & -- & -- & -- 
& \LEFTcircle & \LEFTcircle & \LEFTcircle & \LEFTcircle & \LEFTcircle 
& \LEFTcircle & \LEFTcircle & \LEFTcircle & \LEFTcircle & \LEFTcircle \\

Authentication 
& \Circle & -- & -- & -- & -- 
& \Circle & -- & -- & -- & -- 
& \LEFTcircle & \LEFTcircle & \LEFTcircle & \LEFTcircle & \CIRCLE 
& \Circle & -- & -- & -- & --  \\

DID 
& \Circle & -- & -- & -- & -- 
& \Circle & -- & -- & -- & -- 
& \LEFTcircle & \LEFTcircle & \LEFTcircle & \CIRCLE & \CIRCLE  
& \Circle & -- & -- & -- & --  \\

Audit Trails 
& \Circle & -- & -- & -- & -- 
& \Circle & -- & -- & -- & -- 
& \CIRCLE & \CIRCLE & \CIRCLE & \CIRCLE & \CIRCLE 
& \Circle & -- & -- & -- & -- \\

A2A Messaging 
& \LEFTcircle & \CIRCLE & \CIRCLE & \CIRCLE & \CIRCLE 
& \Circle & -- & -- & -- & -- 
& \LEFTcircle & \CIRCLE & \CIRCLE & \CIRCLE & \LEFTcircle 
& \LEFTcircle & \CIRCLE & \CIRCLE & \CIRCLE & \LEFTcircle \\

Malware Detection 
& \Circle & -- & -- & -- & -- 
& \Circle & -- & -- & -- & -- 
& \Circle & -- & -- & -- & -- 
& \Circle & -- & -- & -- & -- \\

Cryptographic Methods 
& \LEFTcircle & \Circle & \Circle & \LEFTcircle & \LEFTcircle 
& \Circle & -- & -- & -- & -- 
& \LEFTcircle & \Circle & \Circle & \LEFTcircle & \LEFTcircle 
& \Circle & -- & -- & -- & --  \\

\midrule
\multicolumn{21}{l}{\textbf{Trust Environments}} \\
\midrule
Role-Based Access Control 
& \Circle & -- & -- & -- & -- 
& \Circle & -- & -- & -- & -- 
& \CIRCLE & \CIRCLE & \CIRCLE & \CIRCLE & \CIRCLE 
& \LEFTcircle & \CIRCLE & \CIRCLE & \CIRCLE & \CIRCLE \\

Sandbox / Isolation 
& \CIRCLE & \Circle & \Circle & \LEFTcircle & \CIRCLE 
& \Circle & -- & -- & -- & -- 
& \CIRCLE & \CIRCLE & \CIRCLE & \CIRCLE & \CIRCLE 
& \LEFTcircle & \Circle & \Circle & \LEFTcircle & \CIRCLE \\

Scoped Memory 
& \LEFTcircle & \CIRCLE & \CIRCLE & \LEFTcircle & \CIRCLE 
& \Circle & -- & -- & -- & -- 
& \LEFTcircle & \CIRCLE & \CIRCLE & \CIRCLE & \CIRCLE 
& \Circle & -- & -- & -- & -- \\

TEE 
& \CIRCLE & \CIRCLE & \Circle & \Circle & \CIRCLE 
& \Circle & -- & -- & -- & -- 
& \CIRCLE & \CIRCLE & \Circle & \Circle & \CIRCLE 
& \CIRCLE & \CIRCLE & \Circle & \Circle & \CIRCLE \\

\bottomrule
\end{tabular}
}
\end{table*}

Table~\ref{tab:trust_eval_full} evaluates representative approaches to agent trust under the TAN framework. The results show that existing techniques tend to concentrate on specific layers of the system, workflow orchestration, communication protocols, or execution environments, rather than enforcing trust as a global invariant of the agent network. When examined through the four design pillars and operational metrics, each category exhibits clear strengths, yet also yield structural issues. 

\subsection{Single-Agent Alignment: Strong Local Robustness, Weak Compositional Guarantees}

Single-agent alignment methods focus on improving the internal behavior of individual agents~\cite{bai2022constitutional, ouyang2022training}. Within a multi-agent setting, these techniques partially improve compositional robustness: better-aligned agents are less likely to produce harmful outputs, and reinforcement learning or adversarial training can increase resistance to perturbations~\cite{madry2017towards}. Operationally, they are generally efficient and scalable. Most introduce limited additional inference latency and moderate resource overhead, and they scale naturally as more agents are added.

However, their trust assumptions remain local. Alignment does not enforce semantic containment across agents~\cite{park2023generativeagents}. Even if each agent is individually aligned, their interaction can still produce misaligned global states due to divergent interpretations of intent. Accountability is also external to the alignment mechanism since these approaches do not encode provenance into state transitions. Cross-boundary reliability remains heuristic, as internal control loops and planning strategies may reduce looping behavior but do not guarantee bounded execution. In this sense, single-agent alignment exemplifies a bolted-on trust philosophy at the model level: trust is encouraged through external tools or prompting rather than enforced through constraints on the internal transition itself~\cite{brown2020language}. Consequently, safety does not reliably compose in networked environments~\cite{amodei2016concrete}.

\subsection{Multi-Agent Workflow Coordination: Broader Interaction Awareness, Yet Reactive and Non-Deterministic}

Multi-agent workflow coordination techniques explicitly address interaction dynamics. Compared to single-agent alignment, these approaches acknowledge that failures arise from composition. They introduce additional layers of reasoning or oversight to regulate how decisions are aggregated and executed. For example, planner--executor structures impose task decomposition, and supervisors distribute responsibility across roles~\cite{shen2023hugginggpt, yao2023react}. Human-agent interaction can significantly strengthen accountability when approvals and logs attribute actions to identifiable actors~\cite{christiano2017deep}.

Despite these advances, most coordination mechanisms remain reactive. Guardrails operate as monitors layered on top of unconstrained transitions, meaning unsafe states remain reachable if detection fails. Voting and debate increase resource overhead and inference latency, often substantially, and their outcomes depend on probabilistic model reasoning, which somewhat limits determinism~\cite{du2024improving}. While some architectures partially improve compositional robustness or cross-boundary reliability, they do not embed invariants fully into the transition~\cite{perez2023discovering}. As a result, coordination methods extend trust beyond the single-agent level but largely retain a bolted-on character at the workflow layer, improving empirical robustness without eliminating structural vulnerability.

\subsection{Protocol-Centric Trust: Deterministic, but with Semantic Gaps and Variable Overhead}

Protocol-centric trust mechanisms shift trust to the communication and identity layer. Their primary strength lies in determinism: schema validation, identity verification, and cryptographic checks are rule-based rather than probabilistic, resulting in strong consistency ($E_d$) and generally favorable scalability ($E_s$)~\cite{anthropic2024mcp}. Lightweight protocol enforcement, such as structured message validation or signature checks, typically introduces minimal inference latency and resource overhead. However, more advanced cryptographic techniques, such as secure multi-party computation \cite{goldreich1998secure,du2001secure}, zero-knowledge proofs \cite{goldreich1994definitions,sun2021survey}, homomorphic encryption \cite{acar2018survey,brakerski2014leveled,cheon2017homomorphic}, or blockchain-backed verification \cite{sayeed2020smart,gadekallu2022blockchain}, can be computationally and communication intensive, substantially increasing both latency ($E_l$) and resource overhead ($E_r$). Thus, while protocol-centric approaches are structurally deterministic, their efficiency depends on the strength and complexity of the guarantees employed.

Functionally, these mechanisms partially constrain the transition function by restricting who may act and how messages must be formed, thereby strengthening compositional robustness at the syntactic level and improving accountability through traceability. Compared to monitoring-based coordination methods, they move closer to a baked-in paradigm because constraints are embedded directly into interaction rules rather than applied retrospectively. Nevertheless, protocol compliance secures syntax and identity, not semantics. A well-formed and authenticated message can still induce a semantically unsafe state transition. Consequently, although protocol-centric trust offers deterministic structural safeguards, it does not enforce semantic containment, leaving unsafe global trajectories reachable despite correct communication.

\subsection{Trust Environments: Strong Execution Containment, Limited Global Semantics}

These mechanisms offer strong determinism and low operational overhead~\cite{costan2016intel, sabt2015trusted}. Access checks, isolation policies, and enclave verification are typically constant-time operations that scale efficiently. Accountability is strengthened because actions can be attributed to roles or isolated execution contexts. Sandboxing and TEE can also impose resource caps, contributing to cross-boundary reliability by preventing unbounded execution or privilege escalation~\cite{costan2016intel, wahbe1993efficient}. Among existing categories, trust environments represent some of the strongest structural constraints.

Yet their scope is primarily infrastructural. Similar to protocol-centric methods, they prevent unauthorized transitions but do not prevent authorized agents from making semantically harmful decisions. For instance, scoped memory reduces information leakage, and RBAC enforces capability boundaries, but neither guarantees that agents interpret each other’s instructions consistently~\cite{denning1976lattice,sandhu1998role}. These approaches embed constraints at the execution layer, i.e., a more baked-in strategy than monitoring systems, but they do not encode global semantic invariants into the transition. Consequently, they mitigate certain classes of failures without addressing semantic misalignment across agents.

\subsection{Key Takeaways}

The evaluation of existing techniques leads to several key observations:

\begin{itemize}
    \item Strengthening individual agents improves robustness but does not guarantee safe global behavior in multi-agent settings.
    \item Multi-agent orchestration reduces obvious failures but remains probabilistic and bolted-on.
    \item Protocol and environment efficiently enforce syntactic correctness, identity, and execution boundaries, yet do not constrain intent alignment.
    \item Across categories, trust is rarely embedded directly into the transition function. Instead, safeguards are layered on top of unconstrained components.
\end{itemize}

In conclusion, these findings indicate that existing methods address fragments of the trust problem but do not transform trust into a system-level invariant. This structural gap motivates the next step, that is, articulating a blueprint for a TAN framework in which trust is architected from the outset.

\section{Blueprint for TAN}
\label{sec:blueprint}




\begin{figure}
    \centering
    \includegraphics[width=0.85\linewidth]{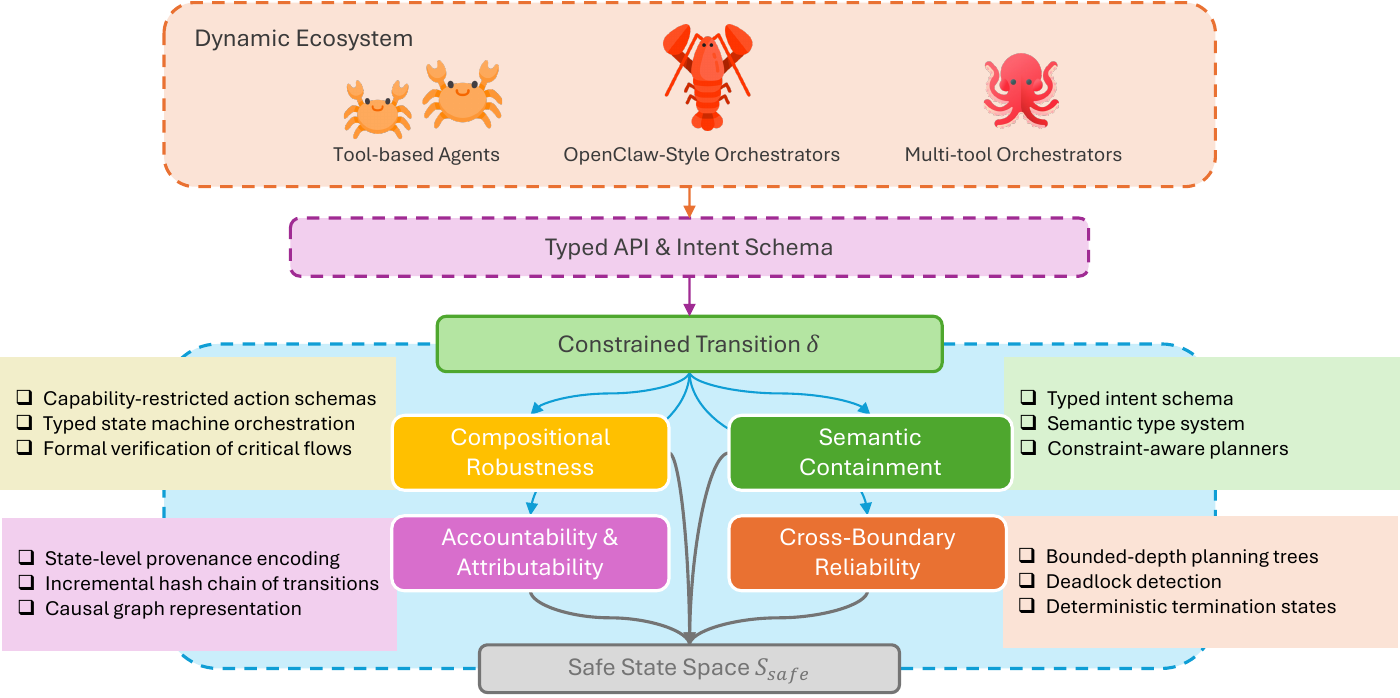}
    \caption{Blueprint for trustworthy agent networks}
    \label{fig:blueprint}
\end{figure}

A blueprint for a TAN must move beyond incremental safeguards and re-architect trust as an intrinsic property of the network. This section outlines how such a shift could address current limitations, what challenges arise in retrofitting existing methods, and which unexplored directions worth further effort.

\subsection{Embedding Compositional Robustness into the Transition Function}

Current alignment and coordination methods assume that safety composes across agents. The evaluation shows this assumption does not hold. Even well-aligned agents, when interacting, can produce unsafe global states. To fix this within existing paradigms would require constraining the transition function $\delta$ itself rather than merely improving individual outputs. Embedding compositional robustness requires combining alignment techniques with capability-restricted action schema, typed interaction protocols, or formally defined state machines that eliminate unsafe transitions by construction. However, imposing such constraints on open-ended LLM agents is challenging because their action spaces are implicitly defined through natural language rather than explicit formal interfaces.

Future TAN architectures must therefore formalize the global state space and allowable transitions more explicitly. Capability-based APIs, typed intent representations, or formally verifiable orchestration layers could restrict reachable states without excessive runtime monitoring. The central challenge is preserving expressivity while embedding constraints: overly rigid transition systems limit agent autonomy, whereas overly flexible systems reintroduce unsafe trajectories. 

\subsection{Achieving Semantic Containment Beyond Syntax}

Protocol-centric and environment-based systems secure syntax and identity but do not guarantee that a receiver’s action preserves the sender’s intended target state within the global state space. Introducing semantic validation layers such as intent classifiers or post-hoc consistency checkers does not alter the underlying transition function $\delta$. Instead, it instantiates the bolted-on verification pattern defined in \Cref{sec:bolted-on}, where safety depends on external monitoring and states outside $\mathcal{S}_{safe}$ remain reachable if monitoring fails.

A TAN blueprint must instead encode semantic constraints directly into interaction contracts. This could involve explicit intent schemas, shared world models with invariant specifications, or constraint-aware planning mechanisms where each state update is validated against formally defined semantic conditions. Another promising direction is the development of semantic type systems that bind permissible actions to intended outcomes. Unlike syntactic validation, semantic containment requires reasoning about intent and consequence, which risks increasing latency and resource cost unless carefully engineered.

\subsection{Intrinsic Accountability Through State-Level Provenance}

Existing systems strengthen accountability through logging, authentication, or audit trails, yet these mechanisms remain external to the state representation. TAN requires that accountability be intrinsic: every state must encode its own causal provenance. Relying on conventional logging mechanisms is insufficient, as logs remain external to the transition dynamics and do not enforce state-level provenance.

A forward-looking blueprint should explore embedding lightweight provenance metadata directly into state updates. Cryptographic commitments, incremental hashing, or structured causal graphs could ensure that any unsafe state can be traced to responsible agents without excessive overhead. However, strong cryptographic guarantees can be computationally heavy, particularly in high-frequency interaction settings. The design challenge is therefore to achieve high determinism and strong attribution while maintaining acceptable latency and resource consumption. Scalable provenance encoding must avoid super-linear complexity as interactions increase.

\subsection{Ensuring Cross-Boundary Reliability and Bounded Execution}

Multi-agent systems are particularly vulnerable to cascading loops and unbounded resource consumption. Existing planner--executor structures, supervisors, sandboxing, and RBAC partially mitigate such risks, yet none guarantee bounded global trajectories. Retrofitting termination guarantees into open-ended language-driven systems is inherently difficult, as generative models are not naturally finite-state.

A TAN architecture should incorporate explicit resource budgeting into the transition function. Each state transition could consume a quantifiable portion of a global resource allowance, ensuring convergence to either a valid target state or a safe termination state. Formal liveness guarantees, bounded-depth planning trees, or checkpoint-based execution contracts could be leveraged. The challenge is balancing reliability with flexibility: strict bounds may constrain problem-solving capability, while loose bounds risk instability. Operationally, such mechanisms must remain scalable and deterministic without introducing excessive runtime overhead.

\subsection{Toward Fully Baked-In Trust}

The cumulative lesson from existing methods is that trust cannot remain an auxiliary layer. Alignment improves individual agents, orchestration regulates workflows, protocols secure communication, and environments constrain execution. However, unsafe states remain reachable because semantic invariants are not embedded into the transition dynamics. Establishing a TAN therefore requires co-designing semantics, protocols, and execution constraints so that safety is enforced within the transition function rather than through post-hoc inspection.

Achieving this objective entails integrating formal state representations, intent-aware interaction contracts, intrinsic provenance encoding, and bounded execution policies into a unified architecture. These properties impose distinct but complementary constraints on the transition function $\delta$: compositional robustness restricts reachable state topology; semantic containment constrains the meaning and effect of actions; intrinsic accountability attaches causal provenance to state updates; and cross-boundary reliability bounds trajectories in time and resource space. Because these constraints operate on the same transition system, they cannot be introduced independently without reintroducing reachability gaps. A coherent TAN architecture must therefore define $\delta$ such that safety, semantics, attribution, and liveness are enforced as coupled invariants.

The principal design challenge lies in embedding these constraints while maintaining efficiency and scalability. Excessively rigid transition systems risk limiting expressivity, whereas overly permissive dynamics undermine safety guarantees. Trust mechanisms must therefore remain deterministic where feasible, compositional across agents, and computationally tractable in large-scale networks.
\section{Alternative Views}

\subsection{Agent-Level Alignment Is Sufficient}

\textbf{The argument:}  
If individual agents are sufficiently aligned through methods such as RLHF or large-scale supervised fine-tuning, then safety will naturally emerge at the network level. Under this view, the problem reduces to improving model alignment. Some proponents go further and suggest that building a single, highly capable “super-agent” would avoid the complexity of multi-agent interaction altogether.

\textbf{Our rebuttal:}  
This view overlooks several structural realities. First, training feasibility remains a major constraint. It is extremely difficult to train a single agent that achieves state-of-the-art performance across all domains, such as software engineering, legal reasoning, planning, and creative writing. 
In practice, multi-agent systems arise precisely because specialization is necessary.

Second, even a perfectly aligned agent cannot remove computational bottlenecks. A monolithic agent processes tasks sequentially, whereas networked systems allow parallel specialization. As task complexity grows, parallel coordination becomes an architectural requirement rather than a convenience.

Most importantly, safety does not compose automatically. An individually aligned agent can still participate in globally unsafe behavior when interacting with others. A reviewer may approve code that appears locally correct but becomes dangerous in a specific deployment configuration. Alignment at the node level does not resolve semantic conflicts at the system level. For these reasons, agent-level alignment remains insufficient as a foundation for network trust.

\subsection{Measuring Trust Would Solve the Problem}

\textbf{The argument:}  
Trust can be treated as a measurable quantity. By assigning each agent a numerical trust score based on past behavior, the system can grant or restrict privileges using simple thresholds. This makes trust computable and easy to operate.

\textbf{Our rebuttal:}  
Reducing trust to a single scalar obscures its context-dependent nature. Trust is not one-dimensional. An agent may be reliable for data analysis but unsuitable for system administration. Collapsing these distinctions into a single score risks granting authority in domains where competence has never been demonstrated.

Moreover, once trust becomes a score, it becomes an optimization target. Agents or adversaries controlling them can accumulate reputation through low-risk actions and later exploit that accumulated credibility to perform a high-impact harmful action. This behavior exposes a fundamental weakness of threshold-based trust models. Quantification alone does not embed structural guarantees, and it merely shifts the problem into a different form.

\subsection{Ensemble Algorithms Ensure Reliability}

\textbf{The argument:}  
Reliability can be achieved through ensemble reasoning among agents. Mechanisms such as majority voting, debate, or chain-of-thought with cross-examination naturally mitigate errors. If multiple agents evaluate each other’s reasoning, hallucinations and unsafe decisions will be filtered out without the need for architectural constraints.

\textbf{Our rebuttal:}  
This argument relies on an assumption of independence that rarely holds in practice. When multiple agents share the same base model or training data, their errors are often correlated. They may converge on the same plausible but incorrect conclusion so that it leads majority voting ineffective.

Additionally, debate dynamics among language models can exhibit conformity effects. Agents may shift their responses toward perceived consensus or authority rather than independently verifying correctness. Without structural safeguards, such as isolation of reasoning traces or blind evaluation, ensemble mechanisms can reinforce shared biases rather than eliminate them. Ensemble reasoning can improve empirical performance, but it does not guarantee structural robustness or semantic consistency.

\subsection{Existing Cryptographic Protocols Already Solve Trust}

\textbf{The argument:}  
Trust in agent networks can be ensured through established cryptographic mechanisms such as encryption, digital signatures, and public key infrastructures. If agents communicate over secure channels with verified identities, the network should be trustworthy.

\textbf{Our rebuttal:}  
This perspective conflates communication security with semantic trustworthiness. Cryptographic protocols secure the channel and verify the sender’s identity, but they do not evaluate the meaning or consequences of the transmitted message. A message can be encrypted, authenticated, and fully compliant with protocol rules, yet still contain a hallucinated fact, a malicious instruction, or a semantically misaligned directive.

In other words, cryptography secures the pipe, not the payload’s intent. Trust challenges in agent networks arise at the level of meaning and system, not merely at the level of bits and bytes. Ensuring that messages are authentic does not ensure that they are safe in context. Therefore, while cryptographic protocols are essential for secure communication, they do not by themselves establish trustworthy multi-agent behavior.

\section{Conclusion}

The main contribution of this work is to show that trust in Agent-to-Agent systems is fundamentally an architectural problem rather than a purely supervisory one. Existing methods remain largely bolted-on: they improve local robustness through external monitoring, filtering, and repair, but they do not prevent unsafe global behavior from emerging through agent composition. In contrast, we argue that trustworthy agent ecosystems require trust to be baked into the system itself, such that unsafe trajectories are restricted by construction. To formalize this perspective, we introduce the Trustworthy Agent Network (TAN) framework and identify four core principles, compositional robustness, semantic containment, accountability, and cross-boundary reliability, as the foundation for trust at the network level. Future work should convert this conceptual framing into implementable architectures, formal verification targets, and benchmarkable design criteria for real-world multi-agent systems.

\bibliography{refs}
\bibliographystyle{plain}

\end{document}